\documentclass{amsart}
\usepackage{amsaddr}
\usepackage{bm,booktabs,graphicx}

\usepackage{setspace}
\newcommand{\ra}[1]{\renewcommand{\arraystretch}{#1}}

\begin{document}

\title{ABtree: An Algorithm for Subgroup-Based Treatment Assignment}

\author{Derek Feng}
\address{%
Department of Statistics \& Yale Institute for Network Science\\
Yale University, New Haven, CT%
}
\email{derek.feng@yale.edu}
\author{Xiaofei Wang}
\address{%
Department of Mathematics and Statistics \\
Amherst College, Amherst, MA%
}
\email{swang@amherst.edu}


\begin{abstract}
Given two possible treatments, there may exist subgroups who benefit greater from one treatment than the other. This problem is relevant to the field of marketing, where treatments may correspond to different ways of selling a product. It is similarly relevant to the field of public policy, where treatments may correspond to specific government programs. And finally, personalized medicine is a field wholly devoted to understanding which subgroups of individuals will benefit from particular medical treatments. We present a computationally fast tree-based method, \textit{ABtree}, for treatment effect differentiation. Unlike other methods, \emph{ABtree} specifically produces decision rules for optimal treatment assignment on a per-individual basis. The treatment choices are selected for maximizing the overall occurrence of a desired binary outcome, conditional on a set of covariates. In this poster, we present the methodology on tree growth and pruning, and show performance results when applied to simulated data as well as real data.
\end{abstract}

\maketitle

\section{Introduction}

Knowing one's customers is of great importance to small and large businesses alike. The field of market segmentation is devoted to the identification of subgroups of a target audience whose consumption behaviors -- direct mail response rates, purchases, frequency of store visits, types of purchases, and so on -- differ. In many cases, the goal is to predict a categorical or quantitative response variable for individual targets based on various predictors. Tree-based methods are particularly desirable for tackling such problems, because they gracefully handle the problem of variable selection and present outputs in an intuitive way. \cite{Thrasher:1991hq} advocated the use of Classification and Regression Trees (CART), first introduced in \cite{Breiman:1984tu}, for classifying potential customers into one of two classes. \cite{Haughton:1997kz} argued that Chi-Square Automatic Interaction Detection (CHAID), first introduced in \cite{Kass:1980df}, performs equally well and is simpler to use. \cite{Maclachlan:1981jo} showed that the closely related Multivariate Automatic Interaction Detection (MAID) can be used to segment customers when the response variable is quantitative and multivariate. 

With the advent of the internet, the E-commerce boom presents additional opportunities, not only for understanding customers, but also for responding to customer preferences. On a website, it is possible to experiment with two alternate store-fronts, randomly displaying one or the other to incoming visitors, and directly measure success from each. The field of A/B testing addresses the question: when there are two options to choose from, A and B, which one produces the better outcome? Given $n$ visits, each individual visit to a website generates data consisting of $(Y_i, \bm{X_i}, T_i)$, where $\bm{X_i}$ is a multivariate vector of characteristics relating to the user, such as type of web browser, operating system, and time of visit, $T_i \in \{A,B\}$ is the version of the website displayed to the user, and $Y_i$ corresponds to some profit measure. Profit measures can be quantitative -- such as visit duration or number of ad clicks -- or categorical -- such as whether the user made a purchase. The standard problem involves determining the single version $T'\in \{A,B\}$ that maximizes expected profit across all future users. Specifically, 
\[
  T' = \operatorname{argmax}_{t\in\{A,B\}}\sum_{n+1}^\infty \mathbb{E}(Y_i|T_i=t).
\]

One drawback of this approach is that it inherently ignores individual characteristics $\bm{X_i}$. For this reason, it is worth reformulating A/B testing as a market segmentation problem; rather than choosing a single $T'$ to be applied to all future customers, differences in customer characteristics should lead to different preferences, and therefore a meaningful objective is to arrive at a series of decision rules for $T_i'$ tailored to the $i$-th customer (with characteristic vector $\bm{x_i}$):

\begin{equation}
    T_i' = \operatorname{argmax}_{t_i\in\{A,B\}}\mathbb{E}(Y_i|\bm{X_i} = \bm{x_i}, T_i=t_i),\quad i>n. \label{eq:intro_profit}
\end{equation}

In this paper, we present a novel method, \textit{ABtree}, that yields a decision tree for determining $T_i'$. \emph{ABtree} assumes $Y_i\in\{0,1\}$ and covariates $X$ are quantitative, categorical, or both. We begin by reviewing relevant literature in related fields, followed by describing our methodology in detail, fully specifying our algorithm with the splitting and pruning procedures. We then evaluate the performance of our method on a diverse set of simulated A/B testing data. We also show the results of \emph{ABtree} on a real-world employment assistance dataset. We finally conclude with a discussion of the extensibility of A/B testing to other problems.

\section{Related Work}

We find parallels in recent advances in personalized medicine, wherein a chief concern is determining subgroups of patients who would benefit from a particular medical treatment. In this setup, $\bm{X}$ are the patient characteristics, such as weight, blood pressure, and cholesterol, and $T$ are the treatments received. The response variable $Y$ in such endeavors is typically far more complex. For example, when dealing with patient outcomes, tree-based methods often need to consider survival models and censoring \cite{Loh:2015eh}. A few approaches assume quantitative $Y$ \cite{Su:2009ty,Dusseldorp:2014dv,laber2015tree}. We briefly review the approaches that consider categorical $Y$. The Virtual Twins (VT) method takes a two-step approach, first using random forests to predict probabilities of success for each individual assuming both treatments, and then using the differences in those probabilities as a response variable in CART \cite{Foster:2011jr}. Another method, Subgroup Identification based on Differential Effect Search (SIDES), yields regions in the covariate space that are likely to have different effect sizes via repeated hypothesis testing \cite{Lipkovich:2011kk}. Both SIDES and VT target the exploration of possible subgroup structures but neither proposes the assignment of treatments.

The area of treatment effect heterogeneity has received considerable and continued attention from myriad sources \cite{angrist2004treatment,imai2013estimating,wager2015estimation}. In particular, \cite{imai2013estimating} proposes an SVM-based method for modeling treatment effects while considering multi-way interactions between the treatment and covariates. We note that while treatment effect heterogeneity is a driving force in treatment selection, we are not interested in the magnitude of treatment effects per se, but rather in an optimal assignment of treatments for the utilitarian goal of profit maximization. None of these methods address this objective.


\section{Methodology}

Formally, suppose we had access to $n$ training examples $\left\{ (Y_i, \bm{X_i}, T_i)\,:\, i = 1, \ldots, n \right\}$, where the \textbf{profit} $Y_i \in \left\{ 0,1 \right\}$ is the response variable, $\bm{X_i} = (X_{i1}, \ldots, X_{ip}) \in \mathcal{X}$ is the associated $p$-dimensional covariate vector, and $T_i \in \left\{ A,B \right\}$ indicates the treatment received. 


An individualized treatment rule is a function $\pi: \mathcal{X} \rightarrow \left\{ A,B \right\}$ that, given a customer with $\bm{X} = \bm{x}$, yields a treatment $\pi(\bm{x})$ for that customer. Our goal is to find the optimal choice of treatment rule $\pi^{\star}$ -- that is, the $\pi$ that maximizes the expected profit,
\begin{equation}
  \pi^{\star} = \arg\max_{\pi}\mathbb{E}\left[ Y | \bm{X} = \bm{x}, T = \pi(\bm{x}) \right]. \label{eqn:pi_star}
\end{equation}
This optimization is intractable though, so we instead adopt the heuristics of decision trees, and consider a data-driven approach to finding a good approximation to $\pi^{\star}$. Hence, we seek a $\hat\pi$
that is piecewise constant on recursive binary partitions of the covariate space, and is comparable to $\pi^{\star}$ in expected profit.
That is, $\hat\pi$ estimates a constant $\hat \pi(\bm{x})$ to all $\bm{x} \in S_j$. For conciseness, we define $\hat \pi(S_j):=\hat \pi(\bm{x})$ for $\bm{x}\in S_j$, which yields the following simplified expected profit for any individual with $\bm{x}\in S_j$,
\begin{align*}
  P(\bm{x}, \hat\pi(S_j)) := \mathbb{E}\left[ Y | \bm{X} = \bm{x} , T = \hat \pi(S_j) \right].
\end{align*}
Given the $n$ training examples $\left\{ (Y_i, \bm{X_i}, T_i)\,:\, i = 1, \ldots, n \right\}$, an empirical counterpart to this expectation is
\begin{equation}
  \tilde P(S_j, \hat\pi(S_j)) :=
  \frac{1}{|\left\{i: \bm{x_i} \in S_j, T_i = \hat \pi (S_j)\right\}|}
  \sum_{i: \bm{x_i} \in S_j, T_i = \hat \pi (S_j)} y_i, \label{eqn:r_tilde}
\end{equation}
which is the empirical mean profit for data points where the covariate is in the subspace $S_j$, and the treatment applied is $\hat \pi (S_j)$. Thus, the optimal choice of $\hat{\pi}(S_j)$ to maximize \eqref{eqn:r_tilde} has a simple solution: pick $\hat \pi (S_j)$ to be the treatment $t \in \left\{ A,B \right\}$ with the larger $\tilde P (S_j, t)$.

More generally, by aggregating across all subspaces, the optimal choice of $\hat{\pi}$ is
\begin{align}
  &\max_{\pi} \sum_{j} |S_j| \cdot \tilde P (S_j, \pi(S_j)) \nonumber
  \\=&\sum_{j} |S_j| \max \left\{ \tilde P (S_j, A), \tilde P (S_j, B) \right\} \label{eqn:bigmax}
\end{align}
It will be helpful later to define the summand in \eqref{eqn:bigmax}:
\begin{align*}
  Q(S_j) := |S_j| \max \left\{ \tilde P (S_j, A), \tilde P (S_j, B) \right\}.
\end{align*}





The mechanism for determining the set of regions $\bm{S}$ is described in the next subsections. For the most part, we follow the conventions laid out in CART \cite{Breiman:1984tu}: we build an initial tree, prune it, and then select the best tree via a holdout set (alternatively, through cross-validation). Thus, to fully delineate our algorithm, it suffices to describe the splitting criteria at each node, the stopping conditions for the tree growth, and the pruning technique.

 
\subsection{Splitting Criteria}

A split is induced by a threshold $\tau$ on the $k$-th predictor of $\bm{X_i}$, which we denote $X_{ik}$. If $X_{ik}$ is continuous, then the split corresponds to the binary question $X_{ik} \leq \tau$. 
When $X_{ik}$ is categorical ($V$ defining the set of all possible categories), the split corresponds to the binary question $X_{ik} \in U \subseteq V$. While this poses computational issues when the search is made over the power set of $V$, there are ways to circumvent such computational barriers. The general technique is to order the categories according to the mean response within each category, and then proceed as if $X_{ik}$ were ordinal.
Another approach is to simply consider splits that correspond to the binary question $X_{ik} = \tau$. We adopt the latter approach, as the implementational advantages outweigh the sometimes marginal gains in tree performance.

Suppose we are in the subspace $S \subseteq \mathcal{X}$. Let us fix a choice of predictor $k$, and assume that $X_{ik}$ is continuous for now. Then, the goal is to pick a $\tau$ such that the subspaces induced by the split are maximal with respect to \eqref{eqn:bigmax}. In other words, we are solving the following optimization:
\begin{equation}
  \max_{\tau} \left\{
    Q(L_{\tau, k} \cap S) + Q(R_{\tau, k} \cap S)
  \right\}, \label{eqn:prfmax}
\end{equation}
where $L_{\tau, k} = \left\{X_i: X_{ik} \leq \tau \right\}, R_{\tau, k} = \left\{X_i: X_{ik} > \tau \right\}$.
Note that $\tilde P(S, T)$ involves only two simple quantities,
\begin{align*}
  |\left\{i: \bm{x_i} \in S, T_i = T\right\}| &=: n_{S}^{T}, \\
  \sum_{i: \bm{x_i} \in S, T_i = T} y_i &=: y_{S}^{T}.
\end{align*}
Thus, \eqref{eqn:prfmax} can be rewritten as
\begin{equation}
  \max_{\tau}
  \left\{ \max\left\{
      \frac{y_{S \cap L_{\tau,k}}^{A}}{n_{S \cap L_{\tau,k}}^{A}},
      \frac{y_{S \cap L_{\tau,k}}^{B}}{n_{S \cap L_{\tau,k}}^{B}},
    \right\} + 
    \max\left\{
      \frac{y_{S \cap R_{\tau,k}}^{A}}{n_{S \cap R_{\tau,k}}^{A}},
      \frac{y_{S \cap R_{\tau,k}}^{B}}{n_{S \cap R_{\tau,k}}^{B}},
    \right\} \right\}
    \label{eqn:bigopt}
\end{equation}
Because the terms in \eqref{eqn:bigopt} depend only on simple counts, this optimization can be solved efficiently. Thus, we iterate over the $p$ predictors and pick the one with the largest value of \eqref{eqn:prfmax}. For categorical predictors, the only difference lies in the definition of $L_{\tau, k}, R_{\tau, k}$.





\subsection{Tree Growth}

To ensure that the comparisons performed in \eqref{eqn:bigopt} are fair ones, and we have a good representation of both treatments in every node of the tree, we adopt similar stopping conditions to those found in standard decision tree methods. There is a minimum split parameter, which is the minimum number of observations needed from both treatment groups in a node in order for a split to be considered. Relative to standard methods, our parameter effectively doubles the number of observations needed in a node. Similarly, our minimum bucket parameter -- the minimum number of observations in a terminal node -- applies simultaneously for both treatments. This also ensures that the quantities in \eqref{eqn:bigopt} are well defined.


\subsection{Pruning}

We follow the pruning technique proposed by \cite{Breiman:1984tu}, whereby a sequence of optimal subtrees is formed by iteratively removing the weakest link of the tree. In our case, this corresponds to the pair of leaf nodes having a common parent that produce the smallest increase in \eqref{eqn:prfmax}.

Having formed a sequence of trees, the optimal tree can be chosen either using a hold-out set, or by cross-validation. Given the nature of the applications for our algorithm, where data is by no means limited in quantity, we recommend using a hold-out set.
Unfortunately, as this is an \emph{unsupervised} learning task, there is no clear measure for the performance of a tree.
Our proxy for the performance is the fraction of treatment assignments predicted by the tree that match the assignment in the hold-out set.

\section{Simulations} 
\label{sec:simulations}
In this section, we evaluate \emph{ABtree}'s treatment selection results to those of (1) a random treatment allocation and (2) the existing A/B test treatment allocation. We measure performance using a mean profit score. In the next subsections, we begin by describing the simulation data and the comparison procedure, with a detailed review of the methods under consideration. We then discuss our specific scoring method before showing the actual simulation results.

We implemented our simulations in \texttt{R}. An \texttt{R} package \texttt{ABtree} is under development.

\subsection{Simulation Data} 
\label{sub:simulation_data}

We consider generative models for our simulated data. Tables~\ref{tbl:simparam1} and \ref{tbl:simparam2} summarize the parameters used. To mimic the conditions of A/B testing, we assign $T$ to be either 0 or 1 with equal probability. The covariates are generated independently from $T$. $\phi_k$ fully specifies the relationship between the probability of success $p$, and a set of covariates $\bm{X}$ and choice of treatment $T$. We can then simulate from Bernoulli($p$) to determine the outcome $Y$.

\begin{table}[t]
\centering
\caption{Simulation Setup Overview}
\ra{1.3}
\begin{tabular}{@{}lll@{}} 
\toprule
Variable & Notation & Distribution\\
\midrule
Treatment & $T$ & Bernoulli$(0.5)$\\
Covariates & $\left\{X_j\right\}_{j=1}^5$ & Uniform($0,1$)\\
Logit($p$) & $log\left(\frac{p}{1-p}\right)$ & $\phi_k(\bm{X}, T)$\\
Profit & $Y$ & Bernoulli$\left(\frac{e^p}{1+e^p}\right)$\\
\bottomrule\label{tbl:simparam1}
\end{tabular}
\end{table}

\subsection{Procedure} 
\label{sub:procedure}
For each $\phi_k$ setting, we simulate 50 datasets. Each dataset has 5000 rows, five covariates, one treatment, and one response, the profit. We divide the dataset into three parts: 50\% training set, 25\% validation set, and 25\% test set. For each method, we proceed in the following way on a given dataset:
\begin{enumerate}
  \item \textbf{Model-based treatment assignment}.
        Use the model to generate treatment choices $T'$ on the test set.
  \item \textbf{Counterfactual simulation}. Simulate $\tilde Y$ using $\phi(\bm{X}, T')$.
  \item \textbf{Evaluation}. Calculate mean profit $\bar P(\tilde Y)$.
\end{enumerate}
Step 1 will vary in complexity depending on the method used. We discuss steps 2 and 3 in greater detail in the next subsection.

The four methods under consideration appear below.

\begin{enumerate}
  \item \textbf{Random assignment}. Treatments are assigned at random to all individuals regardless of $\bm{X}$. The model is simply Bernoulli($0.5$).
  \item \textbf{A/B testing}. A single treatment is assigned to all $T_i'=T'$. Without loss of generality, we assume the control and treatment options are $A$ and $B$ respectively. We run a one-sided hypothesis test on the combined training and validation set to determine if there is evidence at the significance level of 0.05 that the treatment yields a higher average profit. If so, we assign $T'=B$; otherwise, we assign $T'=A$. 
  \item \textbf{ABtree (no pruning)}. We run \emph{ABtree} on the combined training and validation set. In fitting the model only the first four covariates $\{X_j\}_{j=1}^4$ are used. For some $\phi_k$, the fifth $X_5$ is important in determining the treatment effect. We exclude $X_5$ from modeling to mimic real world examples where true drivers of treatment effects may not be measured. We then generate treatment choices $T_i'$ for each individual in the test set.
  \item \textbf{ABtree}. We use the same covariates as in the previous approach. We run \emph{ABtree} on the training set and prune the resulting tree using the validation set. We then generate treatment choices $T_i'$ for each individual in the test set.

\end{enumerate}

\subsection{Performance Evaluation} 
\label{sub:performance_score}
For either choice of treatment $T_i'$ assigned to observation $(Y_i, \bm{X_i}, T_i)$ under regime $k$ in the test set, we can simulate a counterfactual profit $\tilde Y_i$ via Bernoulli($logit^{-1}(\phi_k(\bm{X_i}, T_i'))$). Consistent with the objective of A/B testing, we score each method on basis of mean counterfactual profit, defined as:
\begin{align*}
\bar P(\tilde Y) := \frac{1}{n}\tilde Y
\end{align*}

Intuitively, a method that assigns treatment choices resulting in higher average profits is preferred. Therefore, we equate $\bar P(\tilde Y)$ with better performance.
We note that such a performance score is only meaningful in simulation settings where treatment effects are pre-specified. In real data examples, counterfactual profits cannot be calculated and therefore accurate comparisons against other methods are impossible.

\subsection{Simulation Results} 
\label{sub:simulation_results}

\begin{table}[t]
\centering
\caption{Response Function Summary}
\ra{1.3}
\begin{tabular}{@{}ll@{}} 
\toprule
$k$ & $\phi_k(\bm{X}, T)$\\
\midrule
1 & $2 T \cdot \operatorname{sgn}(X_{i1} - 0.2) + X_{i3} + X_{i4}$\\
2 & $2 T \cdot \operatorname{sgn}(X_{i1}) \cdot \operatorname{sgn}(X_{i2} - 0.3) + X_{i2} + 0.2 X_{i3} + 0.5 X_{i4}$\\
3 & $3 T \cdot \operatorname{sgn}(X_{i1}) + 2 X_{i2} + X_{i3} + 0.5 X_{i5}$\\
4 & $3 T \cdot \operatorname{sgn}(X_{i1}) + T \cdot \operatorname{sgn}(X_{i2}) + X_{i3}$ \\
\bottomrule\label{tbl:simparam2}
\end{tabular}
\end{table}

\begin{figure*}
\centering
\includegraphics[width=\textwidth]{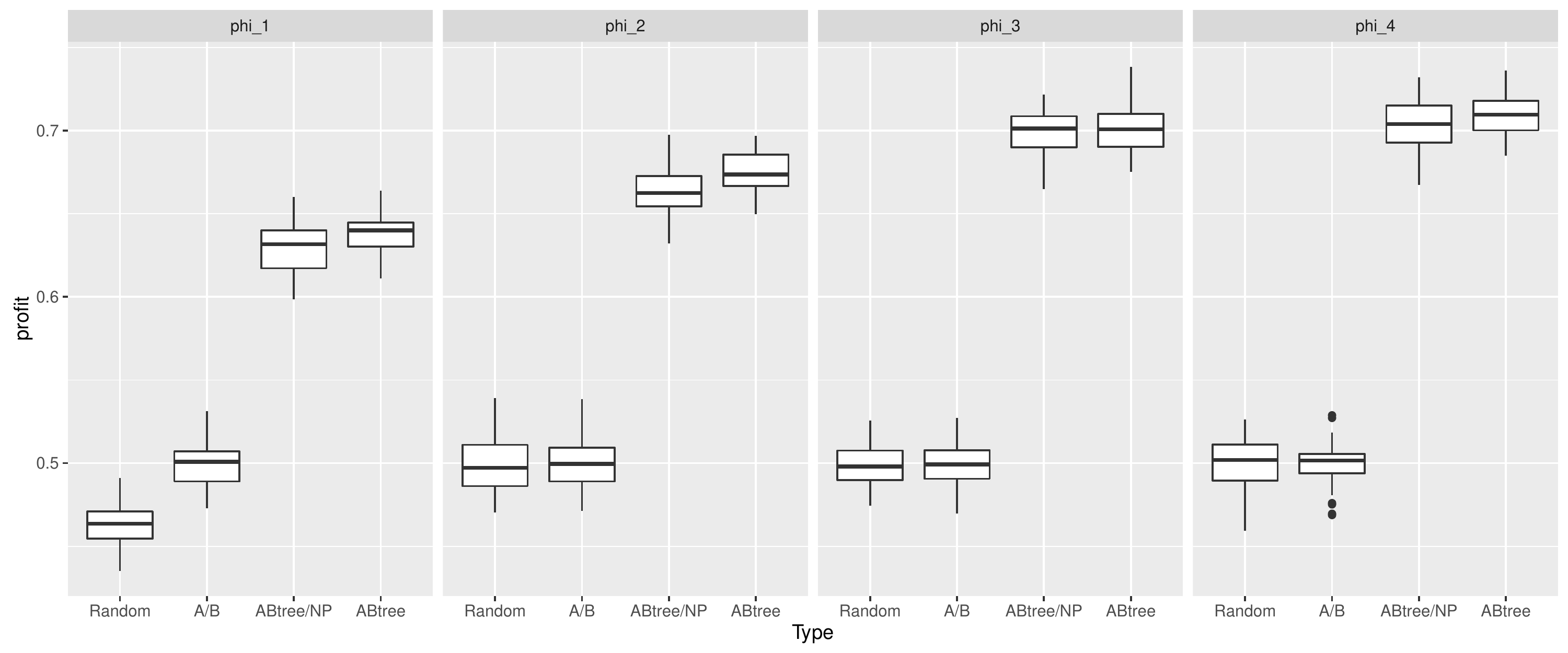}
\caption{Boxplot of mean profit across the different methods and different response functions $\phi_k$}\label{fig:boxplot}
\end{figure*}

A boxplot of the results is shown in Figure~\ref{fig:boxplot}. It is immediately clear from the figure that our algorithm comfortably beats both random assignment and the standard procedure in A/B testing.
This means \emph{ABtree} is successfully capturing the treatment effects found in the $\phi_{k}$. The performance is comparable across the different $\phi_k$, though $\phi_1$ and $\phi_2$ are noticably lower than $\phi_3$ and $\phi_4$, owing in part to the increased signal of the treatment effect (3 vs 2 in the coefficient).
It is worth pointing out that \emph{ABtree} is able to handle the existence of external noise (as in $\phi_3$).
On the other hand, the A/B testing procedure fails to beat random assignment for all cases other than $\phi_1$.

An interesting observation is that the improvements due to pruning are very minor at best. Given the computational cost associated with pruning, it might be prudent in certain applications to forgo the pruning step in lieu of using a (considerably) larger sample size.

\section{Application: National Supported Work Demonstration} 
\label{sec:application_national_supported_work_study}
Due to the proprietary nature of A/B testing data, we are unable to provide an example of \emph{ABtree} applied to an actual A/B testing dataset. We instead examine a National Supported Work Demonstrated dataset \cite{LaLonde:1986bq}, subsequently explored in \cite{imai2013estimating}. The National Supported Work Demonstration
(NSW) was a large-scale national and private
program designed to provide work experience for disadvantaged
workers in the hopes of improving employability. A randomized experiment was conducted in the mid-1970s, in that qualified individuals were randomly allocated to a treatment group, receiving the full benefits of the program, and a control group, receiving nothing. The outcome of interest is a binary indicator of whether individual earnings increased during the term of the experiment. We refer interested readers to \cite{LaLonde:1986bq} for full details.

\begin{figure}
\centering
\includegraphics[width=0.5\textwidth]{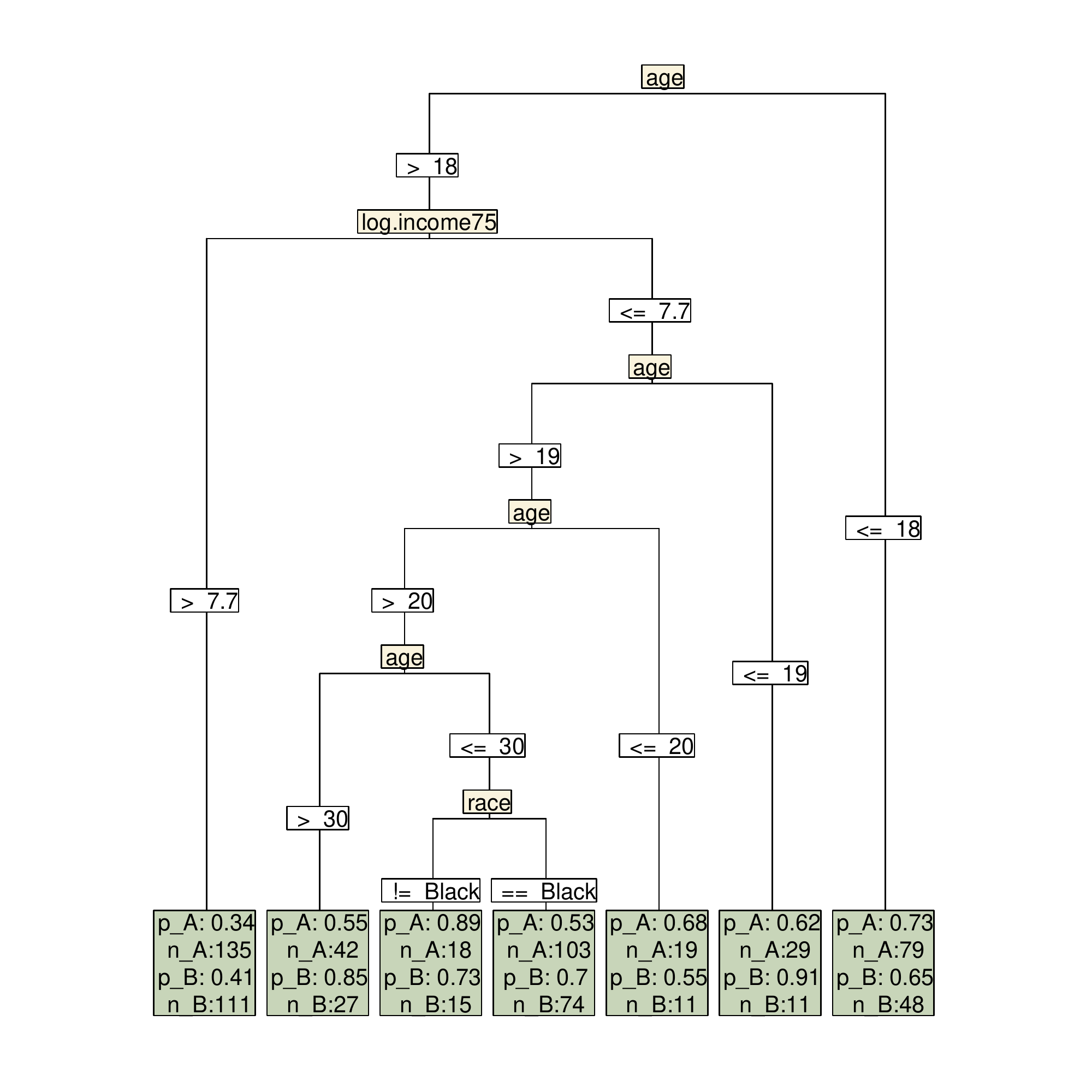}
\caption{NSW segmentation results via \emph{ABtree}. Treatment A is the control (no benefits) and treatment B is the provision of assistance under the program. The yellow boxes in the leaf nodes summarize the proportion of success and sample size under each treatment.}\label{fig:nsw}
\end{figure}

We ran \emph{ABtree} to model whether individual earnings increased conditional on treatment group with covariates of subject-specific information recorded at the start of the study in 1975; these include quantitative variables of age, years of education, and log of income,  as well as categorical variables of race, marital status, attainment of high school degree, and unemployment status. Figure~\ref{fig:nsw} shows the results, starting with an initial split by age above and below 18. The tree suggests that individuals younger than or equal to 18 would not benefit from the NSW program. Among those older than 18 years of age, the sample is further divided by log of earnings in 1975. Most interestingly, despite the wealth of information passed to the algorithm, age is the most important covariate for deciding whether an individual will benefit from the program. 

We note that we cannot compare our results with those in \cite{imai2013estimating} because their method targets estimating treatment effect heterogeneity on a per-individual basis whereas ours is aimed at assigning treatments to maximize the occurrence of a desired outcome. 

\section{Discussion} 
\label{sec:discussion}
In this paper, we presented a novel method, \emph{ABtree}, for identifying subgroups of individuals that respond more positively to one treatment versus another. \emph{ABtree} was motivated by the A/B testing problem; in simulations, we showed that \emph{ABtree} is superior to the existing A/B testing solution that completely ignores individual characteristics. For contextual purposes, we opted to focus on the case where our response variable is binary. However, our results extend naturally with minimal modifications to the continuous case.

In addition to its ability to take a decision-oriented approach on analyzing treatment effects, \emph{ABtree} has many core strengths. For one, as a tree-based method, it has the advantage of yielding highly interpretable decision rules that are easy to understand. The degree of interpretability can be adjusted by increasing or decreasing the maximum tree depth. In cases where interpretability is not a chief consideration, ensemble learning approaches such as bagging and random forests can be directly applied with \emph{ABtree}, which helps to reduce variance and produces smooth decision boundaries. Secondly, the proposed treatment assignment step is entirely apparent given the \emph{ABtree} structure; differences in profit within the terminal leaves are exposed in a plot of the final, preferably pruned, tree. Finally, \emph{ABtree} has no additional tuning parameters other than the standard tree algorithm parameters such as maximum depth size, minimum node size for considering a split, and minimum node size in a leaf, and runs remarkably fast.

In addition to use in A/B testing, we also showed that \emph{ABtree} produces intuitive and credible results on a real-world dataset from the NSW Demonstration study. In fact, there are numerous disciplines in which it is desirable to identify subgroups of individuals for whom a particular treatment is beneficial. In policymaking, for example, it is important to identify subgroups of individuals that will benefit most from a particular subsidy, so that the rules for qualifying for the subsidy can be used to optimize for efficacy. In personalized medicine, as another example, it is important to identify subgroups of individuals that will benefit from a particular drug; \emph{ABtree} can assist with identifying such individuals.

\section{Acknowledgments}
We would like to thank Sahand N. Negahban for helpful suggestions. 
\bibliographystyle{abbrv}
\bibliography{refs,others}

\end{document}